\documentclass[letterpaper, 10 pt, conference]{ieeeconf}  

\IEEEoverridecommandlockouts                              

\usepackage{amsmath} 
\usepackage{amssymb}  
\usepackage{units}
\usepackage{graphicx}
\usepackage{subcaption}	
\usepackage{adjustbox}
\usepackage[font=footnotesize, format=plain, labelfont=bf]{caption}
\usepackage{hyperref}
\usepackage{physics}
\usepackage{todonotes}
\usepackage{cleveref}
\usepackage[dvipsnames]{xcolor}

\usepackage[nocompress]{cite}      

\newif\ifanonymous

\DeclareRobustCommand{\anontext}[2]{%
  \ifanonymous
    \textcolor{orange}{[#2]}%
  \else
    #1%
  \fi
}

\newcommand{\checkmarkgreen}{\textcolor{Green}{\checkmark}}
\newcommand{\crossred}{\textcolor{BrickRed}{$\cross$}}

\begin{document}

\title{\LARGE \bf 
Collaborative Task and Path Planning for Heterogeneous Robotic Teams using Multi-Agent PPO
}
\author{\anontext{Matthias Rubio$^{1}$, Julia Richter$^{1}$, Hendrik Kolvenbach$^{1}$, and Marco Hutter$^{1}$}{Authors}%
\thanks{$^{1}$\anontext{Robotic Systems Lab (RSL), ETH Zürich, Zürich, Switzerland}{Affiliation}}%
}

\maketitle
\thispagestyle{empty}
\pagestyle{empty}

\begin{abstract}

Efficient robotic extraterrestrial exploration requires robots with diverse capabilities, ranging from scientific measurement tools to advanced locomotion.
A robotic team enables the distribution of tasks over multiple specialized subsystems, each providing specific expertise to complete the mission. 
The central challenge lies in efficiently coordinating the team to maximize utilization and the extraction of scientific value. Classical planning algorithms scale poorly with problem size, leading to long planning cycles and high inference costs due to the combinatorial growth of possible robot-target allocations and possible trajectories. 
Learning-based methods are a viable alternative that move the scaling concern from runtime to training time, setting a critical step towards achieving real-time planning. 
In this work, we present a collaborative planning strategy based on Multi-Agent Proximal Policy Optimization (MAPPO) to coordinate a team of heterogeneous robots to solve a complex target allocation and scheduling problem. We benchmark our approach against single-objective optimal solutions obtained through exhaustive search and evaluate its ability to perform online replanning in the context of a planetary exploration scenario.


\end{abstract}

\section{Introduction}

Unmanned surface exploration may require not only different locomotion techniques to overcome harsh terrain but also a variety of scientific equipment and specific devices for physical interaction with the environment. However, a single robot has only a limited capacity. Therefore, distributing task-specific components across a team of specialized robots allows multiple tasks to be performed in parallel, reducing overall mission time \cite{Arm_comp_2023}. 

In 2021, the \textit{Perseverance} rover deployed the helicopter \textit{Ingenuity} on Mars, achieving the first powered flight on another planet \cite{Balaram2021}. This groundbreaking mission proved that the use of multiple agents with alternative locomotion techniques has the potential to enhance future planetary exploration missions. 
Similarly, Arm et al. used a team of legged robots with complementary skills to perform an Earth-based resource prospecting study \cite{Arm_science_2023}. The team was able of performing different tasks in a short time frame, showing that collaboration increases efficiency compared to single-robot exploration \cite{Arm_comp_2023}.

However, in the latter example, the task allocation and task sequence were determined manually by five human operators. As this becomes more complicated for more robots and tasks, especially under the communication delays and constraints of planetary missions, manual planning needs to be replaced by an algorithm that coordinates the team on a global scale.

\begin{figure}[!t] 
    \centering
    \includegraphics[width=1.0\columnwidth]{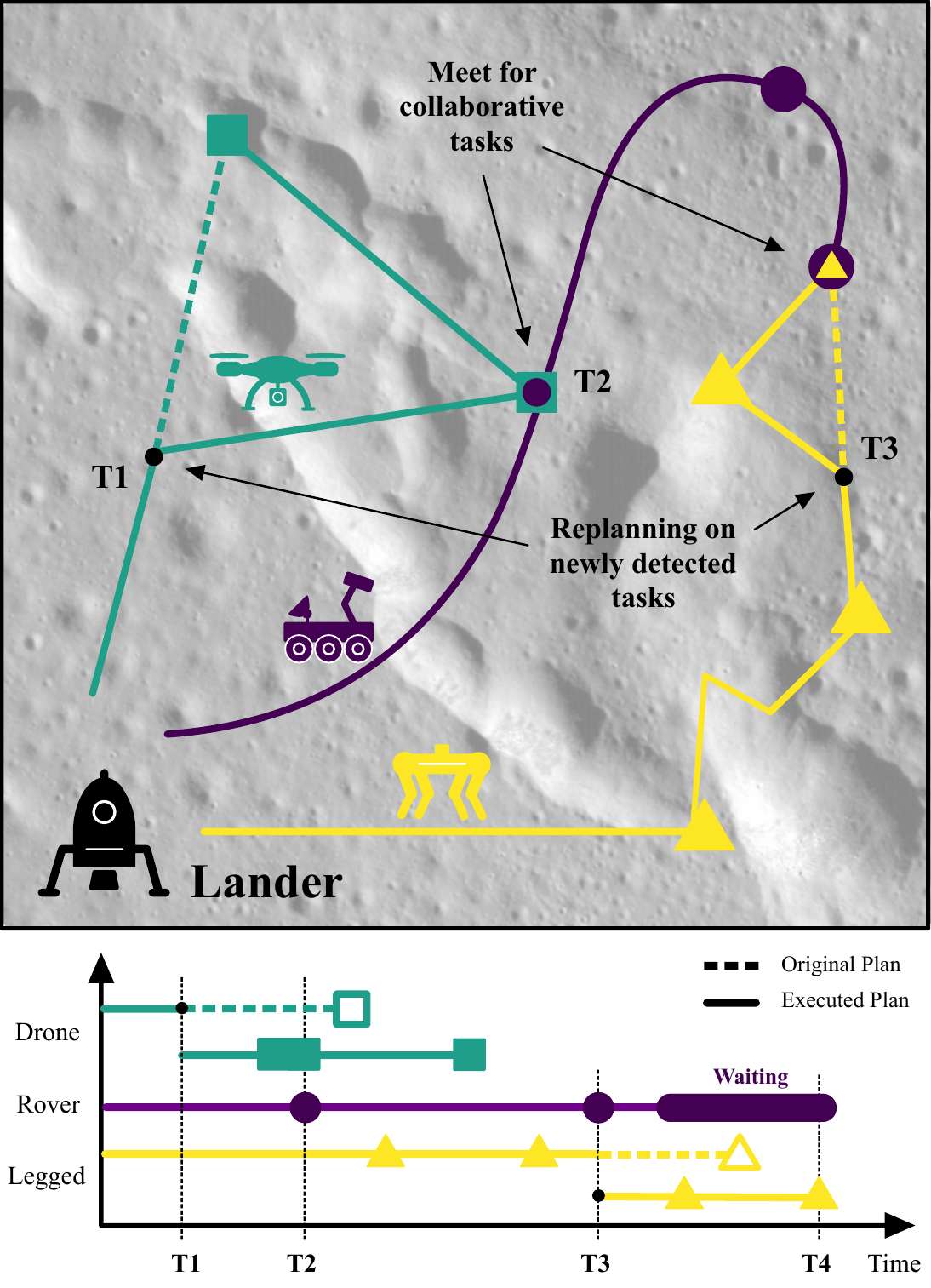}
    \caption{An illustrative plan for a collaborative robot fleet with different specializations, such as flying, walking, or driving. During the mission the drone and the legged robot find new tasks and replan to minimize mission time.}
    \label{fig:map_example}
\end{figure}

\subsection{Challenges of Multi-Robot Planning}

For single-robot path planning, there are many possible algorithmic approaches which Sánchez-Ibáñez et al. systematically categorize \cite{SanchezIbanez2021}. Graph search algorithms, such as A*, make up a well-known subgroup in addition to sampling-based methods. Richter et al. show how to use a multi-objective A* global path planning algorithm for an exploration mission on the moon \cite{Richter2023}. 

However, single-robot path planning algorithms fail to address the additional complexities in multi-agent scenarios. Besides finding an efficient path, the algorithm also needs to \textbf{allocate} a subset of tasks and determine a specific execution order for each robot. Since multiple robots can interact and collaborate, the algorithm additionally needs to \textbf{schedule} the robots appropriately to avoid potential conflicts or exploit synergies between them.

In a space exploration mission, not all the information is known a priori, and new regions of interest, terrain changes, or robot failures lead to unexpected situations. Hence, a fully autonomous robot team should be able to \textbf{replan} and adapt on site according to the available and incoming information. Especially for exploration on another planet, this prevents unnecessary communication delays and idle times while extending the available time for scientific investigation. Due to limited computational resources, such an algorithm also has to adhere to the tight real-time constraints posed by space systems.

Prior work has framed the problem in different ways, e.g., as a multi-traveling salesman problem (MTSP), and taken first steps towards learning-based solutions. Nevertheless, existing methods are limited to assigning agents to targets based on individual path costs and then scheduling them in a separate step, making it difficult to handle both aspects within a single solver. In contrast, our work introduces a fully learning-based method that unifies path planning, task allocation, and scheduling for heterogeneous multi-robot teams. The main contributions can be summarized as follows.

\begin{itemize}
    \item A MAPPO-based reinforcement learning framework for cooperative multi-agent path planning, task allocation, and scheduling.
    \item Benchmark against an optimal exhaustive search baseline, demonstrating competitive performance with improved scalability.
    \item Validation of fast replanning in dynamic environments, highlighting applicability to space exploration missions.
    \item Open-source release of the learning framework.\footnote{{\tt\small \anontext{\url{https://github.com/leggedrobotics/multi\_robot\_global\_planner}}{link}}}
\end{itemize}

The remainder of this paper is structured as follows. In \Cref{sec:method}, our method is thoroughly explained, including a formal description of the problem and the training strategy. The implementation and results on performance and replanning capabilities are presented in \Cref{sec:results}, and finally the limitations of the method and recommendations for future work are discussed in \Cref{sec:discussion}.

\section{Related Work}

\subsection{MTSP and Algorithmic Approaches}

A well-known formulation of path planning problems that includes target allocation is the MTSP, where its numerous variations are listed in \cite{Cheikhrouhou2021}. \textbf{Exact methods} to solve the MTSP include constraint programming \cite{Vali2017ACP} and integer programming algorithms \cite{Sundar2017}. The problem can also be extended to require collision-free paths, such as in \cite{Ma2016}, where Ma et al. present a conflict-based min-cost-flow algorithm on a time-expanded graph, or in \cite{Turpin2013}, where Turpin et al. show how to decouple the target assignment and scheduling problem to solve those parts sequentially. In general, exact methods are able to find a globally optimal solution. However, they tend to scale poorly with the number of agents and targets and therefore result in long computation times, which can go up to hours, as stated in \cite{Cheikhrouhou2021}. This issue makes it hard to perform continuous replanning on constrained computational resources, such as on a space exploration rover.

The state-of-the-art methods in the literature to solve MTSP problems are \textbf{metaheuristic algorithms}, such as NSGA-II \cite{Shuai2019} or particle-swarm optimization techniques, e.g., ant colony optimization \cite{Pamosoaji2020} or the artificial bee colony algorithm \cite{Dong2019, Pandiri2018}. Metaheuristic algorithms are popular because they can be implemented in a computationally efficient way. However, they do not guarantee finding an optimal solution in finite time or even a valid solution at all. Since the problem still scales with the number of agents and targets, a trade-off between computation time and solution quality is inevitable. Jiang et al. address this using their multi-agent planning framework, which consists of two different algorithms that exhibit the trade-offs between plan quality and computational efficiency \cite{Jiang2019}.

Recent work tries to investigate the capabilities of \textbf{learning-based approaches}, where the scaling problem shifts from runtime to training time, enabling the efficient use of computational resources in advance and having almost constant inference times. This can be particularly interesting for space applications that require real-time computation. The standard MTSP is solved using a reinforcement learning approach in \cite{Hu2020} and \cite{Guo2024}, where the problem is separated into two stages. The first stage is a graph neural network that learns how to allocate targets to agents, and the second stage consists of a standard single-traveling salesman problem solver that is also used to supervise the network learning process. This method has been shown to outperform metaheuristic methods in solution quality from small-scale problems up to tens of agents and hundreds of cities. However, to the best of our knowledge, there are no full learning-based approaches trying to extend the MTSP to a case with collaborative tasks, which would include appropriate scheduling of the robots.

\subsection{Learning Collaborative Strategies}
\label{sec:follow_strategy}

To include the full range of subproblems (target allocation, scheduling, and replanning), the problem can be framed within the context of a collaborative game-theoretic approach. In this context, rather than seeking the optimal solution in a large search space, the focus shifts to identifying a general strategy that can be learned empirically and followed consistently by the agents. This motivates the question of whether learning-based methods can be employed to learn such strategies and to what extent they approach optimality.

An unsupervised approach to multi-agent path planning is presented in \cite{Sartoretti_2019} (\textit{PRIMAL}) and extended in \cite{Zhiyao_2020} (\textit{PRIMALc}). The proposed algorithms learn a decentralized strategy to navigate towards targets while avoiding collisions. This concept was also explored in the context of socially aware navigation for multi-robot teams \cite{Wang2023}. 

\noindent They demonstrated how robots can be trained to navigate efficiently in a crowded environment while ensuring safety and comfort for humans around the robots. The learning architecture consists of a spatial-temporal graph neural network that can compute an embedding expressing the human-robot and robot-robot interactions, and an underlying multi-agent proximal policy optimization (MAPPO) algorithm (\cite{Yu2021, Lowe2017}) that learns how to compute the actions for each robot. According to \cite{Yu2021}, MAPPO has been demonstrated in different baseline environments such as the multi-agent particle-world environments \textit{(MPE)}, the \textit{StarCraft} multi-agent challenge, \textit{Google Research Football} and the \textit{Hanabi} challenge. It is therefore a competitive baseline in cooperative multi-agent learning and provides a promising foundation for applying to the target allocation and scheduling problem in an exploration task.

\section{Method}
\label{sec:method}

To learn a strategy that solves the planning problem, we construct a virtual environment containing agents and targets. Agents have the freedom to choose their actions at each step with the goal of solving all targets.

\begin{figure*}[!t] 
    \centering
    \includegraphics[width=\textwidth]{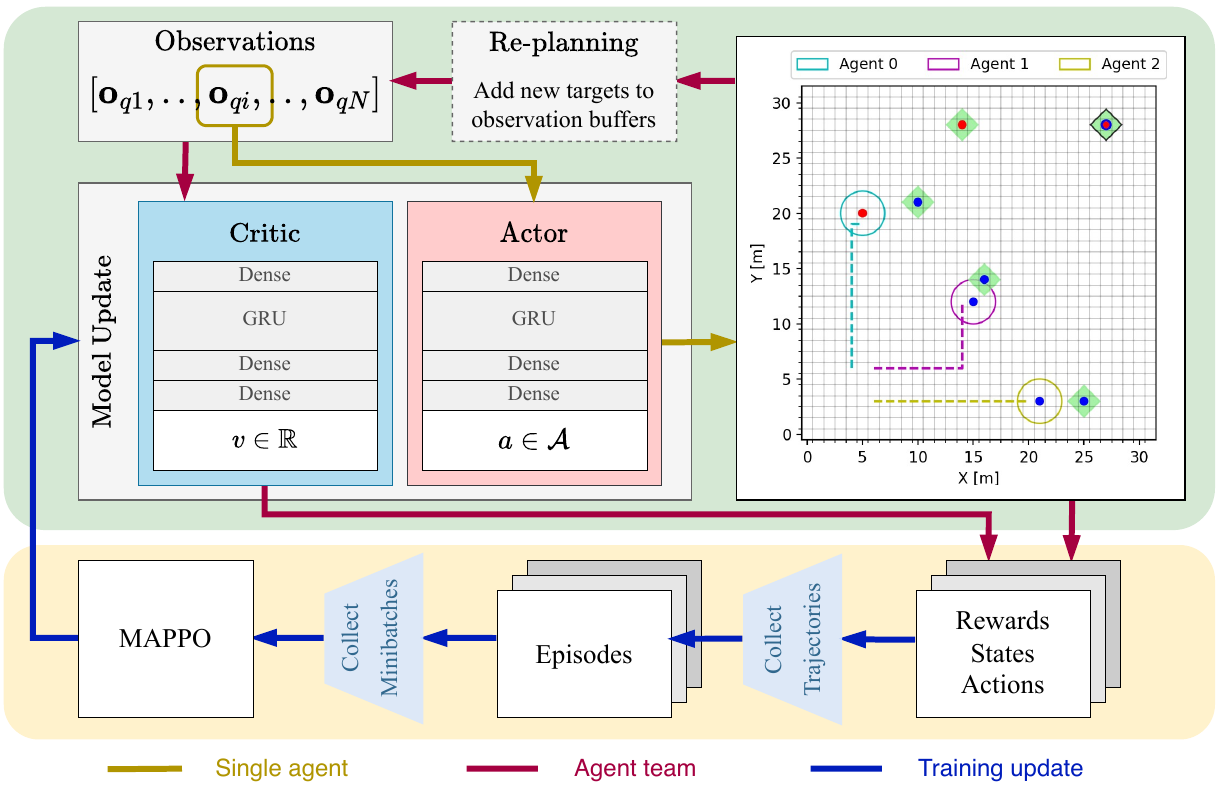}
    \caption{This figure illustrates the complete workflow, highlighting both the execution (green) and training (yellow) phases. The execution block details the network architectures and the placement of the replanning step. The training block shows the MAPPO update sequence. The colored arrows differentiate data flow, specifying whether it applies to all agents, a single agent, or represents aggregated data for training. Furthermore, the environment is visualized as a grid, including the agents and targets with their provided or required skills (colored dots). The targets are marked with a green square, which can have a black border indicating a collaborative target (AND type).}
    \label{fig:architecture_process}
\end{figure*}

\subsection{Training Environment}
\label{sec:training_environment}

The training environment is modeled as a discrete 2D grid (\Cref{fig:architecture_process}) on which the agents $q \in \mathcal{Q}$, $|\mathcal{Q}| = N$ can move using one of the actions $\mathcal{A} = \{up, down, left, right, stay\}$. Each target $t \in \mathcal{T}$, $|\mathcal{T}| = M$ requires a set of skills $\mathrm{S} \in \mathcal{P}(\mathcal{S})$, where $\mathcal{P}(\mathcal{S})$ is the power set of all the skills sets. The targets must be solved by either one of the required skills (OR type) or by all of them (AND type). Each agent provides a set of skills that can be used to solve a target. A target $t \in \mathcal{T}_S(j)$ is solved if, for a time step $j$, all the required skills of the target are in the same position as the target itself. Agents complete an environment if the number of unsolved targets is zero, that is, $| \mathcal{T}_U(j) | = 0$. Furthermore, the action space stays the same throughout the map, even at the border. If an agent attempts to move beyond the boundaries of the map, it will be reverted to the last valid cell it occupied, and the invalid action will be ignored.

\subsection{Observations}
\label{sec:observations}

First, the positions of the targets and all agents are included in the observation of an agent as shown in \Cref{eq:obs_pos}. The symbol $\vb{p} \in \mathbb{N}^2_0$ denotes an absolute position on the grid and $\vb{r}_q^t \in \mathbb{Z}^2$ is the relative position between an agent $q$ and a target $t$. The subscripts $i$ and $k$ are the indices of the agents and targets, and $N = |\mathcal{Q}|$ and $M = |\mathcal{T}|$ are the number of agents or targets, respectively. 
\begin{gather}
    \label{eq:obs_pos_g}
    g(\vb{r}_{q}^t) =  
    \begin{cases}
        \vb{r}_{q}^t &\quad \text{if $t \in \mathcal{T}_U(j)$} \\
        [0,0] &\quad \text{if $t \in \mathcal{T}_S(j)$}
    \end{cases} \\
    \label{eq:obs_pos}
    \vb{o}^{pos}_{qi} = [ \vb{p}_{qi}, ... , \vb{p}_{qN}, g(\vb{r}_{qi}^{t1}), ..., g(\vb{r}_{qi}^{tM})]
\end{gather}
To indicate that a target is solved, its relative position is automatically set to zero by the function $g(\vb{r}_t)$ in \Cref{eq:obs_pos_g} for all agents. The observation structure is kept agent-specific, denoted by the subscript $qi$, always having the position of the current agent in the first entry of the array. Enforcing this structure allows the actor network to associate the observing agent with its position.

In order to solve the targets efficiently, agents need information about the skill sets of the other agents and the skill demands of each target. Since multiple skills can be associated with an agent or a target, the observations in \Cref{eq:obs_skill} are encoded as skill sets written as $\mathrm{S}_{qi}$ or $\mathrm{S}_{tk}$. In practice, skill sets are enumerated and assigned to integer values by a predetermined function $f(\mathrm{S}): \mathcal{P}(\mathcal{S}) \rightarrow \mathbb{N}_0$. 
\begin{equation}
\label{eq:obs_skill}
    \vb{o}^{skill}_{qi} = [ f(\mathrm{S}_{qi}), ..., f(\mathrm{S}_{qN}), f(\mathrm{S}_{t1}), ..., f(\mathrm{S}_{tM}) ]
\end{equation}

Finally, targets must be distinguished by their type (AND or OR type) to indicate whether collaboration is necessary. In \Cref{eq:obs_goal_types}, the types $h$ are encoded as 1 (AND type) or 0 (OR type).
\begin{equation}
\label{eq:obs_goal_types}
    \vb{o}^{goalType} = [ h_{t1}, ..., h_{tM} ]
\end{equation}

The complete observation for an agent $q_i$ is a concatenation of the previously mentioned observations and can be written as:
\begin{equation}
    \vb{o}_{qi} = [ \vb{o}^{pos}_{qi}, \vb{o}^{skill}_{qi}, \vb{o}^{goalType} ]
\end{equation} 

Note that the observation of an agent depends on the number of agents and targets that exist in the environment. Although this design choice prevents variation in numbers during training, it eliminates the need for a separate encoding algorithm or network to handle dynamically sized observations.

\subsection{Rewards}
In the following, the different reward terms used in the training are discussed. To help the agents navigate towards the sparsely distributed targets, an attraction reward (AR) is introduced, which is spread around a target with an increasing value towards the center. The agent receives this reward at each time step, if the target is unsolved and the agent has at least one matching skill with the target, which can be expressed as the condition $P = (t \in \mathcal{T}_U(j)) \wedge (| \mathrm{S}_{qi} \cap \mathrm{S}_{t} |) > 0$. The reward function for one target is shown in \Cref{eq:rew_attr_inner_func}, where $\vb{r}_{t}$ is the relative distance from agent $qi$ to target $t$ and $C^{AR}$ is a constant parameter that determines the spread of the attraction reward. The attraction reward in \Cref{eq:rew_attr_outer_func} is averaged over all the targets to be solved in the environment and normalized with respect to the maximum number of steps $T_{max}$.
\begin{gather}
    h_{qi}(t) = 
    \begin{cases}
    \label{eq:rew_attr_inner_func}
        \exp (-C_{AR} \cdot \norm{\vb{r}_{t}}_2) &\quad \text{if $P$}\\
        0 &\quad \text{otherwise}
    \end{cases} \\
    \label{eq:rew_attr_outer_func}
    r^{AR}_{qi} = \frac{1}{M \cdot T_{max}} \sum_{t \in \mathcal{T}} {h_{qi}(t)}
\end{gather}

Once a target is reached and solved, there is a fixed payout for all agents, regardless of whether they contributed to solving the target. This prevents competition among agents. 

\noindent The target reward (TR) function is shown in \Cref{eq:rew_task} where $j$ refers to the current time step. If all the targets are solved, the collected rewards sum up to 1.
\begin{equation}
\label{eq:rew_task}
    r^{TR}_{qi} = 
    \begin{cases}
        \frac{1}{M} &\quad \text{if $t \in \mathcal{T}_U(j-1) \wedge t \in \mathcal{T}_S(j)$} \\
        0 &\quad \text{otherwise}
    \end{cases}
\end{equation}

To give more importance to skills, a fixed penalty (WC) is added if agents step on a target that does not have a common skill, as shown in \Cref{eq:rew_on_other}.
\begin{equation}
\label{eq:rew_on_other}
    r^{WC}_{qi} = \sum_{t \in \mathcal{T}_U}
    \begin{cases}
        -1 &\quad \text{if $(\norm{\vb{r}_t}_2 = 0) \wedge (| \mathrm{S}_{qi} \cap \mathrm{S}_{t} | = 0)$} \\
        0 &\quad \text{otherwise}
    \end{cases}
\end{equation}

To ensure efficient completion of targets, specifically minimizing the number of required steps, agents are subjected to a nominal cost per movement (SC), which can be formalized as shown in \Cref{eq:rew_step}. The action that leads to the current state is written as $u(j-1)$.
\begin{equation}
\label{eq:rew_step}
    r^{SC}_{qi} = 
    \begin{cases}
        0 &\quad \text{if $u(j-1) = stay$} \\
        -1 &\quad \text{otherwise}
    \end{cases}
\end{equation}

In addition to a minimal number of steps, agents are expected to complete the targets in a minimum time frame, which is introduced as shown in \Cref{eq:rew_solve_time}. The solve-time cost (TC) decreases with the number of solved goals and is normalized by the number of total targets $M$ and the maximum number of steps $T_{max}$ in the episode. It follows that if the agents solve all the targets during an episode, the cost vanishes.
\begin{equation}
\label{eq:rew_solve_time}
    r^{TC}_{qi} = \frac{|\mathcal{T}_U(j)|}{M \cdot T_{max}}
\end{equation} 

Finally, the terminal bonus incentivizes agents to complete an environment further by rewarding them for solving all targets, as shown in \Cref{eq:rew_final_bonus}.
\begin{equation}
\label{eq:rew_final_bonus}
    r^{TB}_{qi} = 
    \begin{cases}
        1 &\quad \text{if $(\mathcal{T}_S(j-1) \subset \mathcal{T}) \wedge (\mathcal{T} \subseteq \mathcal{T}_S(j))$} \\
        0 &\quad \text{otherwise}
    \end{cases}
\end{equation}

The complete reward for an agent computed at each time step is shown in \Cref{eq:full_rew} where each reward term is weighted by a constant parameter $w \in \mathbb{R}$, which allows the relative influence of the rewards to be adjusted in training.
\begin{equation*}
    \vb{r}_{qi} = [r^{AR}_{qi}, r^{TR}_{qi}, r^{WC}_{qi}, r^{SC}_{qi}, r^{TC}_{qi}, r^{TB}_{qi}]^\top
\end{equation*}
\begin{equation*}
    \vb{w} = [w^{AR}, w^{TR}, w^{WC}, w^{SC}, w^{TC}, w^{TB}]
\end{equation*}
\begin{equation}
    \label{eq:full_rew}
    r^{full}_{qi} = \vb{w} \cdot \vb{r}_{qi}
\end{equation}

\subsection{Learning Architecture}
\label{sec:learning_architecture}

As stated in \Cref{sec:follow_strategy}, we use the MAPPO algorithm \cite{Yu2021, Lowe2017}, due to its proven performance in cooperative multi-agent settings. We follow Lowe et al. \cite{Lowe2017}, using an actor-critic structure for the MAPPO networks. The actor and critic networks have a gated recurrent unit (GRU), wrapped by dense layers, as shown in \Cref{fig:architecture_process}. The critic learns a joint value function in a centralized way by taking a concatenation of all the agent observations. On the other hand, the actor execution is decentralized and outputs a probability distribution over the action space. In our case, the actor learns a strategy for an agent with arbitrary skills that can be applied to any agent on the team.

\subsection{Training Strategy}
\label{sec:training_strategy}

The step and solve time costs are initially discouraging. This can create an exploration bottleneck early in training, where agents fail to discover rewarding behaviors and instead converge to a degenerate policy that minimizes penalty by remaining stationary. To overcome this initial difficulty, the training is divided into two parts, the so-called \textbf{bootstrap} and \textbf{refinement}. In bootstrap, agents learn how to navigate to targets based on their skills. In refinement, they learn how to solve the targets efficiently. The rewards are activated as listed in \Cref{tab:reward_activation}.

\begingroup
\setlength{\tabcolsep}{10pt} 
\renewcommand{\arraystretch}{1.5} 
\begin{table}
    \centering
    \begin{tabular}{c|c|c|c|c|c|c}
        Training & AR & TR & WC & SC & TC & BR \\
         \hline
        Bootstrap & \checkmarkgreen & \checkmarkgreen & \checkmarkgreen & \crossred & \crossred & \crossred \\
        Refinement & \checkmarkgreen & \checkmarkgreen & \checkmarkgreen & \checkmarkgreen & \checkmarkgreen & \checkmarkgreen \\
    \end{tabular}
    \caption{Activated rewards during training. AR: Attraction Reward, TR: Target Reward, WC: Wrong Target Cost, SC: Step Cost, TC: Solve Time Cost, BR: Bonus Reward}
    \label{tab:reward_activation}
\end{table}
\endgroup

During refinement training, the attraction reward is maintained to help agents finalize their navigation strategy. Its significantly lower value, compared to other reward terms, is designed to mitigate its effect towards the end of training. In all training runs, the initial positions and the skill sets for the agents and targets, as well as the target types, are randomized. However, to ensure that an environment is solvable, the agent team is always given at least one of each required skill.

\subsection{Implementation}
\label{sec:implementation}

Our pipeline was implemented using the JaxMARL framework \cite{flair2023jaxmarl}, which provides baseline training algorithms and template environments for multi-agent reinforcement learning. The number of environment steps, the number of mini-batches, the number of parallel environments, and the total training steps were manually tuned with respect to the quality of the solution and the GPU hardware constraints. The parameters used for the different policies (with varying numbers of targets) are listed in \Cref{tab:train_params}.
The reward weights for the policies are shown in \Cref{tab:reward_weights}. Depending on which training phase, the weights were set to zero according to the activation rules in \Cref{tab:reward_activation}.

\section{Results}
\label{sec:results}

The main results were obtained by training three agents with two different skills acting on a 32x32 map. Agents had two possible skills, implying three possible skill sets. Targets requiring both skills could be of type AND or OR. 

To analyze how well the method scales with the number of targets, we evaluated three different policies, $\Pi_{5T}, \Pi_{6T}, \Pi_{7T}$, with agents solving five to seven targets. Due to the fixed observation size, as mentioned in \Cref{sec:observations}, the three policies had to be trained separately. During training, actions were sampled from a weighted probability distribution computed by the actor network to allow for a certain amount of exploration. For the evaluation, the action with the maximal probability was applied at each time step.

The actor-critic networks, with approximately 245'000 kernel parameters, were trained on an Nvidia GeForce RTX 4090 GPU. Inference times for the trained network and the baseline were measured on a MacBook Pro with an i5 @ 2.3 GHz with 8GB of RAM.
\begingroup
\setlength{\tabcolsep}{5pt} 
\renewcommand{\arraystretch}{1.5} 
\begin{table}
    \centering
    \begin{tabular}{l|c|c|c}
         & $\Pi_{5T}$ & $\Pi_{6T}$ & $\Pi_{7T}$ \\
        \hline
        $n_{ENV}$ & 16'384 / 30'720 & 16'384 / 26'880 & 16'384 / 24576 \\
        $n_{STEPS}$ & 128 & 128 & 128 \\
        $n_{TRAIN}$ & 572 / 1271 & 572 / 2034 & 1668 / 3814 \\
        $n_{MINI}$ & 16 / 32 & 16 / 28 & 16 / 32 \\
        $Seed$ & 2 / 4 & 2 & 2 \\
    \end{tabular}
    \caption{Modified MAPPO training parameters in top-down order are \textit{environments trained in parallel} / \textit{max steps per environment}, \textit{epochs} / \textit{batch size} / \textit{random generator seed}.}
    \label{tab:train_params}
\end{table}
\endgroup

\begingroup
\setlength{\tabcolsep}{5pt} 
\renewcommand{\arraystretch}{1.5} 
\begin{table}
    \centering
    \begin{tabular}{c|ccccccc}
     & $C_{AR}$ & $w^{AR}$ & $w^{TR}$ & $w^{WC}$ & $w^{SC}$ & $w^{TC}$ & $w^{TB}$ \\
     \hline
     $\Pi_{5T}$ & 0.0075 & 1.0 & 1.0 & 0.25 & 0.3 & 0.5 & 0.2 \\
     $\Pi_{6T}$ & 0.0075 & 1.0 & 1.0 & 0.25 & 0.3 & 0.5 & 0.2 \\
     $\Pi_{7T}$ & 0.04 & 1.0 & 1.0 & 0.25 & 0.3 & 0.5 & 0.2 \\
    \end{tabular}
    \caption{Reward weights for the baseline policy.}
    \label{tab:reward_weights}
\end{table}
\endgroup

\subsection{Evaluation Metrics}
\label{sec:evaluation_criteria}

We introduce a set of metrics to quantify the overall performance of our method. In the results, the metrics will be averaged over a series of randomly generated environments to obtain a statistical evaluation and marked with a bar $\overline{M}$ accordingly.

\subsubsection{Success Rate}
The success rate represents the number of solved environments $K_{solved}$ versus the total number of simulated environments $K_{sims}$. In solved environments, all targets were solved.
\begin{equation}
    M_{success} = \frac{K_{solved}}{K_{sims}}
\end{equation}

\subsubsection{Solve Time}
$T_{solved}$ is the number of time steps until the agents have solved all the targets, and $T_{max}$ is equal to the maximum number of environment steps $n_{STEPS}$ as listed in \Cref{tab:train_params}. Therefore, the difference in the metric $M_{st}$ represents how much time is left before the environment is marked as unsolved. We use the difference to allow for a relative comparison with the baseline algorithm.
\begin{equation}
    M_{st} = T_{max} - T_{solved}
\end{equation}

\subsubsection{Total Team Effort}
The total team effort is defined as the sum of all agent movements. Similarly to the solve time, we define the metric $M_{tte}$ as the negative counterpart, which is the sum of movements that agents have left after all targets were solved. The function $r^{step}$ was defined in \Cref{sec:training_environment} and evaluates to 1 for each moving action at time step $j$.
\begin{equation}
    M_{tte} = \sum_{q \in \mathcal{A}} \left( T_{max} - \sum_{j = 1}^{T_{solved}} |r^{step}_{qi}(j)| \right)
\end{equation}

\subsection{Policy vs. Optimal Solution}
\label{sec:policy_versus_optimal}

Although our multi-agent reinforcement learning (RL) policy is trained for a multi-objective goal, we compare its performance to optimal solutions found via exhaustive search (ES) in \Cref{tab:policy_quality}. This approach provides a quantitative benchmark, demonstrating how closely our policy's performance on each metric approaches its theoretical best-case scenario. We compute two sets of optimal solutions, optimizing ES1 with respect to $M_{st}$ and ES2 with respect to $M_{tte}$. For each policy, we averaged the metrics over 100 different simulated environments.

The success rate for all policies is greater than 90\%, which shows that most environments are solved. 
Furthermore, the three trained policies achieve greater optimality with respect to total team effort (92\%, 91\%, 84\%) than with respect to the solve time (86\%, 81\%, 73\%).
This indicates that the chosen reward weights led to a policy that favors team effort over solving the targets in minimal time. Furthermore, the numbers reveal a decreasing performance across all metrics as the number of targets increases, which is consistent with the problem's growing complexity.

\begingroup
\setlength{\tabcolsep}{10pt} 
\renewcommand{\arraystretch}{2} 
\begin{table}[]
    \centering
    \begin{tabular}{c|c|c|c}
    
                                  & $\Pi_{5T}$ & $\Pi_{6T}$ & $\Pi_{7T}$ \\ \hline
$\overline{M}_{success}$                         &  0.99  &  0.95  &  0.91  \\ \hline
$\overline{M}_{st}^{RL}$\big/ $\overline{M}_{st}^{ES1}$    & 0.86 & 0.81 & 0.73  \\ \hline
$\overline{M}_{tte}^{RL}$\big/ $\overline{M}_{tte}^{ES2}$ &  0.92  & 0.91  & 0.84 \\

    \end{tabular}
    \caption{Performance comparison between three policies with a different number of targets, and the optimal solutions ES1 w.r.t. solve time (st) and ES2 w.r.t. the total team effort (tte). Optimal performance would be represented by a value of 1.}
    \label{tab:policy_quality}
\end{table}
\endgroup

\subsection{Comparison of Inference Time}
\label{sec:shift_complexity}

A notable advantage of a learning-based method is that the inference time remains constant and is independent of the problem's initial conditions. This is because a single forward pass through the network has a time complexity of $O(1)$. This property enables real-time operation in a resource-constrained environment. In contrast, exact methods, such as the ES approach, exhibit an inference time that scales exponentially with the number of agents, targets, or skills. Furthermore, the inference time of these methods varies significantly based on the initial conditions.

\begin{figure}[!htbp] 
    \centering
    \includegraphics[width=1.0\columnwidth]{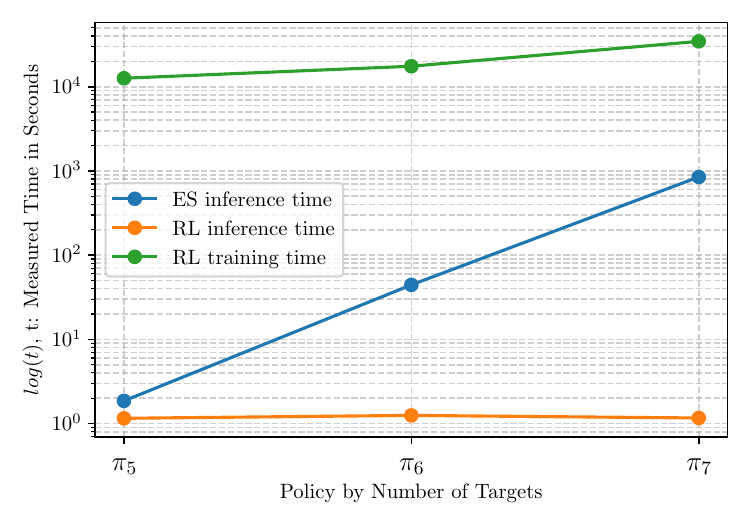}
    \caption{RL inference and training time measurements compared to the inference time of the ES approach with respect to the trained policies by number of solved targets.}
    \label{fig:time_scaling}
\end{figure}

In \Cref{fig:time_scaling}, we show a comparison between the inference times for the ES approach, our policy at runtime, and the corresponding training time for a varying number of targets. The inference time for the RL policy was measured by running a simulation to its limit, which in our case is 128 steps, i.e., forward passes. For both approaches, the time was averaged over 10 simulations. 

The graph reveals an exponential increase of approximately 1.5 orders of magnitude for the ES solving time, whereas the RL training time grows at around 0.25 orders of magnitude. However, the RL training time is orders of magnitude beyond the ES solving time.

\subsection{Replanning Capability}

Having a short and constant inference time for the network opens the possibility of performing online replanning on newly discovered targets. As described in \Cref{sec:observations}, the observation size is fixed and the number of targets cannot be changed for a pretrained network. Thus, we use the observation as a buffer, where new incoming targets replace those that have already been solved.

For this experiment, we used $\Pi_{5T}$ as a baseline and an additional policy $\Pi_{5T5R}$ that was trained such that the agents had to solve the five initial targets and an additional five re-planned targets in an episode. The new targets were randomly generated with different positions, skill sets and goal types and were added to the observation as soon as one of the initial targets had been solved. All training parameters and reward weights were kept the same, as was the training strategy. Both policies were simulated over 1000 environments to solve five initial and five additional targets.

As listed in \Cref{tab:replanning_performance}, the results for both policies are very similar, showing that explicitly training with newly incoming targets does not improve performance. Note that the success rate $M_{success}$ in \Cref{tab:replanning_performance} is lower compared to the results shown in \Cref{sec:policy_versus_optimal}, because an episode was only marked as successful if the agents could solve 10 targets instead of five, while the maximum duration of the episode was still 128.

\begingroup
\setlength{\tabcolsep}{10pt} 
\renewcommand{\arraystretch}{2} 
\begin{table}[]
    \centering
    \begin{tabular}{c|c|c|c}
         & $M_{success}$ & $M_{st}$ & $M_{tte}$ \\
         \hline
        $\Pi_{5T}$ & 84.2 \% & 85.70 ± 20.1  & 232.5 ± 72.8 \\
         \hline
        $\Pi_{5T5R}$ & 84.1 \% & 85.73 ± 20.9 & 220.0 ± 81.3 \\
    \end{tabular}
    \caption{Baseline policy $\Pi_{5T}$ versus a policy $\Pi_{5T5R}$ that was trained including the possibility of obtaining new targets during an episode. Simulated over 1000 random (seed=10) environments with an episode length of 128.}
    \label{tab:replanning_performance}
\end{table}
\endgroup

\section{Discussion}
\label{sec:discussion}

\subsection{Policy Performance}
In \Cref{sec:policy_versus_optimal}, we compared our method against two optimal solutions obtained by ES. The learned policies achieved up to 86\% optimal solution quality with respect to the solve time and up to 92\% with respect to the total team effort. These results demonstrate that a reinforcement learning policy can approximate near-optimal performance with significantly lower computational cost at runtime. The higher optimality with respect to total team effort compared to solve time indicates that the chosen reward weights biased the policies toward minimizing overall effort rather than completion time. In addition, performance across all metrics decreases as the number of targets increases, reflecting the growing complexity of the problem.

Still, the final success rate leaves room for improvement. Training with a longer episode length could have helped to explore more edge cases and increase performance. However, GPU memory limitations imposed a trade-off between episode length (i.e., exploration horizon) and the number of environments that could be trained in parallel.

As is common in reinforcement learning, reward tuning was a central challenge. When increasing the number of targets, we observed that the previously tuned parameters remained applicable only up to a certain number, beyond which re-adjustment of the weights is necessary. For example, the intensity of the attraction reward needed to be reduced when training with more targets to prevent mutual cancellation of target rewards due to excessive overlap.

\subsection{Inference Time}
When the RL method is deployed on real hardware, the inference time of a fixed-size network remains constant for an increasing number of targets. This property can prove advantageous when designing real-time embedded systems. 
In contrast, exact methods scale exponentially with the number of agents, targets, or skills and are highly sensitive to initial conditions.
However, the findings in \Cref{fig:time_scaling} also show that the inherent complexity of the problem is not eliminated, but instead is shifted and condensed to the training phase. Training times and computational demands scale rapidly with the number of targets, agents, and skills, slowing down policy development. Moreover, the corresponding scaling rate in \Cref{sec:shift_complexity} is likely higher, since the performance decreased when additional targets were introduced (\Cref{tab:policy_quality}), suggesting that the training was not fully exploited.

\subsection{Replanning}
By using the observation as a buffer and replacing solved targets with new ones, our method demonstrated its ability to perform online replanning. In real-world missions, this could be applied iteratively, rather than calculating a complete solution upfront. Plans would be generated for shorter horizons and dynamically adjusted as agents observe new targets. Such an iterative process reduces the number of forward passes, thereby improving the method's computational efficiency. 

\subsection{Limitations}
The main limitation of this approach is the fixed observation size of the current architecture, which limits the number of targets as well as the team size, thus restricting scalability. The work of Wang et al. \cite{Wang2023}, which used a graph neural network to learn an embedding for social awareness in a multi-robot team, offers a possible starting point for computing observations that are independent of the number of entities. Another idea is presented by Hafner et al. in \textit{DreamerV3} \cite{hafner2023mastering}, where they showed how to extend an actor-critic approach with an additional auto-encoder to learn a world model representation given an instantaneous partial observation of the agent.

\section{Conclusion}

This work explored a reinforcement learning approach to the multi-agent global path planning and scheduling problem, where agents learn an emergent strategy for team coordination and scheduling to efficiently solve a set of targets on a grid. Using a MAPPO-based centralized training framework, we derived decentralized policies that achieved near-optimal solution quality.

We observed that with a learning-based method, the complexity of the problem is shifted from runtime to training time. This property can be especially interesting for real-time systems with limited onboard compute, as inference requires only constant-time forward passes. Furthermore, the policy demonstrated the ability to online replanning by using the observation as a buffer for newly incoming targets. 

Looking forward, a key step toward generalization and scalability is to design an observation structure that is independent of the number of agents and targets. Such representations would enable the discovery of a general policy applicable to different team sizes, numbers of targets, and skill compositions.

\addtolength{\textheight}{-12cm}   






\bibliographystyle{IEEEtran}
\bibliography{bibtex}

\end{document}